\documentclass[11pt]{article}

\usepackage[margin=1in]{geometry}
\usepackage{amsmath,amssymb}
\usepackage{booktabs}
\usepackage{multirow}
\usepackage{array}
\usepackage{enumitem}
\usepackage{xcolor}
\usepackage[numbers,sort&compress]{natbib}
\usepackage[colorlinks=true,linkcolor=blue,citecolor=blue,urlcolor=blue]{hyperref}

\newcommand{\grounded}{\textsf{grounded}}
\newcommand{\notgrounded}{\textsf{not\_grounded}}
\newcommand{\abstain}{\textsf{abstain}}

\title{More Debate, Same Evidence:\\
Structural Limits of Homogeneous Multi-Agent Groundedness Judges}
\author{Yuelyu Ji}
\date{}

\begin{document}
\maketitle

\begin{abstract}
Large language model (LLM) judges are increasingly organized as multi-agent
panels under the assumption that exchanging critiques improves judgment
quality. We test this assumption for \emph{groundedness verification}, where a
judge must determine whether a claim is supported by the supplied evidence.
We evaluate a homogeneous three-agent panel on six public fact-verification and
hallucination-detection benchmarks. Relative to a fixed single-agent reference,
the panel's system-level accuracy difference ranges from $+8.5$ to $-4.4$
percentage points: two datasets show reliable gains, one shows a reliable loss,
and three are statistically inconclusive. Because the reference and panel use
different model variants, these differences characterize the complete systems
rather than isolate a causal debate effect.

Mechanistic analyses nevertheless reveal a consistent pattern within the
panel. Second-round arguments add little semantic novelty, class-specific
recall shifts explain much of the observed change, and self-reported confidence
is only weakly predictive of correctness. We then use reflective prompt
optimization to test whether persona wording is the bottleneck. A
skeptic-first rewrite improves held-out accuracy by $5.83$ points in one
exploratory comparison and by $5.00$ points in a replication, but the first
result does not survive correction across the seven tested alternatives.
Moreover, $49$ of $120$ held-out items remain wrong under every prompt
configuration. The panel's agents exhibit strongly correlated errors,
especially between the advocate and domain-expert roles.

These results identify a structural limitation of homogeneous debate: agents
that share a model, evidence pool, and fixed aggregation rule can move a
decision boundary without acquiring independent evidence. Reliable gains may
therefore require model or evidence diversity, or direct optimization of the
aggregation and abstention policy, rather than additional rounds or persona
rewriting alone.
\end{abstract}

\section{Introduction}

LLMs are widely used as evaluators of factuality, faithfulness, and response
quality~\citep{liu2023geval,zheng2023judging}. A natural extension replaces one
judge with several role-conditioned agents that exchange arguments before
producing a collective verdict. Multi-agent debate has improved mathematical
and strategic reasoning in prior work~\citep{du2023debate}, and ChatEval applies
a related idea to LLM-based evaluation~\citep{chan2023chateval}. These results
motivate a tempting hypothesis: if a single groundedness judge is uncertain or
biased, a panel of agents may correct it through deliberation.

Groundedness judging is a demanding test of that hypothesis. The task is not to
produce a plausible answer, but to decide whether a claim follows from a fixed
evidence set. When every agent sees the same evidence and is instantiated from
the same model, discussion cannot reveal a source that no agent received. The
panel may still help by surfacing overlooked details, changing the decision
threshold, or reducing idiosyncratic decoding errors. Conversely, shared model
priors may make the agents' errors highly correlated, leaving majority voting
and prompt refinement with little useful diversity.

This paper studies a homogeneous debate panel for groundedness verification.
Three GPT-5.5 Chat agents---a skeptic, an advocate, and a domain expert---read
the same claim and evidence, debate for two rounds, and are combined by a fixed
rule-based aggregator. We evaluate the system on six public benchmarks spanning
Wikipedia entailment, retrieval-augmented generation, QA hallucination, and
scientific claim verification. We then analyze what changes between rounds and
apply GEPA-style reflective prompt evolution~\citep{agrawal2025gepa} to the
three persona prompts.

Our study addresses four questions:
\begin{enumerate}[leftmargin=1.5em]
    \item How consistently does a homogeneous debate panel outperform a
    single-agent reference across groundedness tasks?
    \item Does the second round add substantively new reasoning, or mainly
    shift class-specific decision thresholds?
    \item Do confidence, speaking order, and reasoning-quality metrics explain
    when debate helps?
    \item Can reflective prompt optimization remove the panel's errors, or are
    they constrained by shared-model and fixed-aggregation structure?
\end{enumerate}

The principal finding is not that debate uniformly helps or uniformly fails.
Its apparent benefit is task-dependent, while its mechanism is largely a
recalibration of existing judgments. Prompt optimization finds one useful
skeptic rewrite, but many errors persist across every evaluated configuration.
This combination of low round-to-round novelty, weak confidence calibration,
and correlated persona errors points to structural rather than merely verbal
limitations.

\paragraph{Contributions.}
\begin{itemize}[leftmargin=1.5em]
    \item We provide a six-benchmark evaluation of a homogeneous debate judge,
    reporting paired, cluster-aware uncertainty rather than only aggregate
    accuracy.
    \item We connect accuracy changes to class-specific recall shifts,
    round-to-round semantic novelty, confidence calibration, and persona error
    correlations.
    \item We use reflective prompt optimization as a diagnostic intervention
    and show that prompt rewrites yield limited, selection-sensitive gains with
    no ensemble headroom in the evaluated family.
    \item We separate conclusions supported by the present system comparison
    from causal claims that would require a same-model single-agent control.
\end{itemize}

\section{Related Work}

\paragraph{LLMs as judges.}
LLM-based evaluation offers flexible natural-language criteria and has shown
substantial agreement with human preferences in several settings
~\citep{liu2023geval,zheng2023judging}. At the same time, an LLM judge can
inherit position, verbosity, self-preference, and domain biases from its base
model. Groundedness verification is especially sensitive to plausible but
unsupported inferences: a fluent rationale does not establish that the cited
evidence entails a claim.

\paragraph{Multi-agent debate.}
Debate methods instantiate multiple agents, expose them to one another's
arguments, and aggregate their answers. Du et al.~\citep{du2023debate} report
improvements on reasoning and factuality tasks with multi-round debate.
ChatEval~\citep{chan2023chateval} uses a multi-agent referee team for evaluating
generated text. These studies establish debate as a plausible inference-time
strategy, but do not imply that homogeneous debate will help when all agents
share the same evidence bottleneck. Our focus is not open-ended answer
generation; it is binary support verification under a fixed context.

\paragraph{Groundedness and hallucination benchmarks.}
Our evaluation combines fact verification and hallucination detection.
VitaminC targets sensitivity to subtle factual revisions
~\citep{schuster2021vitaminc}; WiCE provides real-world Wikipedia claims paired
with fine-grained evidence~\citep{kamoi2023wice}; RAGTruth annotates
hallucinations in retrieval-augmented outputs~\citep{niu2024ragtruth}; HaluEval
collects generated and human-annotated hallucination examples
~\citep{li2023halueval}; and SciFact pairs scientific claims with supporting or
refuting abstracts~\citep{wadden2020scifact}. Their differing domains and label
constructions let us test whether a single debate recipe transfers.

\paragraph{Reasoning analysis and prompt optimization.}
ROSCOE provides interpretable metrics for semantic consistency,
informativeness, and related properties of reasoning traces
~\citep{golovneva2022roscoe}. We use these metrics diagnostically rather than
as ground truth for reasoning quality. GEPA uses natural-language reflection
and Pareto selection to evolve prompts in compound LLM systems
~\citep{agrawal2025gepa}. We apply its central idea to the persona prompts and
ask whether prompt-level search can overcome shared panel errors.

\paragraph{Broader application contexts.}
Reliable evidence use and calibrated model decisions are relevant beyond
fact-verification benchmarks. Graph-based recommendation and data-driven
business-intelligence systems increasingly rely on learned representations and
automated decision pipelines~\citep{liu2025research,chen2025design}. In systems
closer to our setting, answer attribution in reasoning models and the robustness
of GraphRAG expose risks when generated conclusions cannot be reliably traced
to retrieved evidence~\citep{wang2025reasoningretrievalstudyanswer,liang2025graphrag}.
Related reliability questions also arise in policy-aligned healthcare AI and
deep neural ECG classification~\citep{chen2026dlsg,song2026deep}, as well as in
multimodal recommendation systems that fuse heterogeneous representations
~\citep{wang2025dlrrec,tian2024mmrec}. We cite these works as motivating
application contexts rather than as groundedness-judging baselines.

\section{Task and Evaluation Data}
\label{sec:data}

Each example contains a claim or candidate response, an evidence context, and a
gold binary label. We map \emph{supported}, \emph{entailed}, and
\emph{faithful} to \grounded{}, and map \emph{refuted}, \emph{contradicted},
\emph{hallucinated}, and not-enough-information labels to
\notgrounded{}. Each evaluation subset is class-balanced. An \abstain{} output
is treated as incorrect under strict binary accuracy; we analyze abstention
separately when studying prompt optimization.

Table~\ref{tab:datasets} summarizes the six datasets. The Composed set is
constructed from WiCE by joining four claims; it is labeled \grounded{} only
when all four subclaims are supported. It tests whether the judge verifies full
coverage rather than accepting a partially supported conjunction.

\begin{table}[t]
\centering
\small
\caption{Evaluation datasets. $n$ is the number of distinct examples. Each
example is evaluated under all six persona speaking orders in the main debate
study.}
\label{tab:datasets}
\begin{tabular}{p{2.2cm}p{7.0cm}r}
\toprule
Dataset & Verification target & $n$ \\
\midrule
VitaminC & Wikipedia fact verification under subtle edits & 250 \\
RAGTruth & Hallucination in retrieval-augmented outputs & 500 \\
WiCE & Fine-grained entailment from Wikipedia evidence & 500 \\
HaluEval & QA-style hallucination recognition & 250 \\
SciFact & Scientific claim verification & 250 \\
Composed & Four-claim WiCE conjunction; all claims must hold & 200 \\
\midrule
Total & & 1,950 \\
\bottomrule
\end{tabular}
\end{table}

\section{A Homogeneous Debate Judge}
\label{sec:method}

\subsection{Panel Roles}

The panel contains three GPT-5.5 Chat agents. All agents receive the same claim
and evidence context, but their prompts impose different verification roles.

\begin{itemize}[leftmargin=1.5em]
    \item \textbf{Skeptic:} precision-oriented falsifier. It checks for
    contradictions, unsupported subclaims, and incomplete evidence. It is the
    only role with an explicit veto path.
    \item \textbf{Advocate:} recall-oriented defender. It accepts support by
    paraphrase, coreference, and direct inference, and rejects mainly on clear
    contradiction or missing support.
    \item \textbf{Domain expert:} coverage auditor. It checks domain-relevant
    entities, quantities, relations, and subclaim coverage.
\end{itemize}

Each agent produces a verdict, a confidence value, and a rationale. In round
one, agents reason independently. In round two, each agent sees the first-round
arguments and may revise its verdict. The complete panel therefore makes six
LLM calls per example.

\subsection{Fixed Aggregation}

A deterministic meta-judge combines the final agent verdicts. It is not an LLM
and is not modified during prompt optimization. Rules execute in the order
shown in Table~\ref{tab:aggregate}. The ordering is important: a two-agent
majority is resolved before the skeptic veto, so the skeptic cannot overturn a
majority \grounded{} decision.

\begin{table}[t]
\centering
\small
\caption{Rule-based aggregation cascade.}
\label{tab:aggregate}
\begin{tabular}{cll}
\toprule
Order & Condition & Output \\
\midrule
1 & Structural short-circuit & Early verdict \\
2 & At least two agents vote \grounded{} & \grounded{} \\
3 & At least two agents vote \notgrounded{} & \notgrounded{} \\
4 & Input is unjudgeable & Unjudgeable \\
5 & Skeptic veto on a split vote & \notgrounded{} \\
6 & Otherwise & \abstain{} \\
\bottomrule
\end{tabular}
\end{table}

\subsection{What the Design Can and Cannot Do}

The agents can draw attention to details already present in the shared context,
counter role-specific biases, and average decoding noise. They cannot add a
missing source or obtain independent model knowledge through discussion. We
therefore distinguish two possible mechanisms:

\begin{description}[leftmargin=1.8em,style=nextline]
    \item[Evidence expansion] A later round introduces substantively new
    evidence or an independent inference unavailable in the first round.
    \item[Recalibration] The same evidence is interpreted under a shifted
    threshold, changing the balance between \grounded{}, \notgrounded{}, and
    \abstain{} without adding information.
\end{description}

Our analyses in Sections~\ref{sec:results} and~\ref{sec:mechanism} test which
description better fits the observed behavior.

\section{Experimental Protocol}
\label{sec:protocol}

\paragraph{Models.}
The three-agent panel uses GPT-5.5 Chat. The fixed single-agent reference in the
six-dataset comparison uses GPT-5 mini. The reference is included to measure
the end-to-end difference between the available systems. Because the model
variants differ, the comparison does \emph{not} identify the causal effect of
adding debate. Within the panel, all roles use the same model and evidence.
The prompt-optimization study holds the GPT-5.5 Chat panel, data, and aggregator
fixed and changes only persona prompts.

\paragraph{Repeated speaking orders.}
For each example, we evaluate all six permutations of skeptic, advocate, and
domain-expert speaking order. Accuracy and recall analyses pool these repeated
trials, while uncertainty is clustered by the underlying example so that the
six orders are not treated as independent observations.

\paragraph{Metrics and tests.}
The primary metric is strict binary accuracy. We additionally report class
recall, the fraction of label flips that become correct, confidence AUROC,
round-to-round semantic novelty, and pairwise persona error correlations.
Accuracy differences use a 1,000-sample cluster bootstrap by example; the GEPA
held-out comparison uses a 10,000-sample paired bootstrap and exact McNemar
tests. We report uncorrected and family-wise interpretations for the seven
prompt alternatives.

\paragraph{Terminology.}
We call the three-agent configuration \emph{homogeneous} because its agents use
the same model. We call Table~\ref{tab:main} a \emph{system-level comparison},
not a debate-only ablation, because its single-agent reference uses a different
model variant.

\section{Cross-Dataset Results}
\label{sec:results}

\subsection{The System-Level Difference Is Task-Dependent}

Table~\ref{tab:main} reports the single-agent reference (R1), the aggregated
two-round panel (R2), and their paired difference. The difference ranges from
$+8.5$ points on VitaminC to $-4.4$ points on SciFact. VitaminC and Composed
show reliable positive differences; SciFact shows a reliable negative
difference. The intervals for RAGTruth, HaluEval, and WiCE include zero.

\begin{table}[t]
\centering
\small
\caption{System-level comparison across six datasets. Confidence intervals use
a cluster bootstrap by example and preserve the six speaking orders. ``ns''
indicates that the interval includes zero. R1 uses GPT-5 mini; R2 is the
homogeneous GPT-5.5 Chat panel, so $\Delta$ is not a debate-only causal effect.}
\label{tab:main}
\begin{tabular}{lcccc}
\toprule
Dataset & R1 & R2 & $\Delta$ & 95\% CI \\
\midrule
VitaminC & 0.768 & 0.853 & $+0.085$ & $[+0.045,+0.124]$ \\
Composed & 0.485 & 0.505 & $+0.020$ & $[+0.008,+0.035]$ \\
RAGTruth & 0.638 & 0.658 & $+0.020$ & $[-0.002,+0.042]$ ns \\
HaluEval & 0.824 & 0.829 & $+0.005$ & $[-0.019,+0.029]$ ns \\
WiCE & 0.828 & 0.809 & $-0.020$ & $[-0.039,+0.000]$ ns \\
SciFact & 0.813 & 0.769 & $-0.044$ & $[-0.069,-0.020]$ \\
\bottomrule
\end{tabular}
\end{table}

The result does not support a universal ``debate helps'' claim. It also does
not support a universal failure claim: the VitaminC difference is sizeable and
the Composed difference is small but stable. Instead, the system appears to
interact with dataset-specific class difficulty and domain conventions.

\subsection{Label Flips Are Not Uniformly Corrective}

Among examples whose label changes between the reference and panel, the
fraction ending on the correct label is $54.3\%$ for RAGTruth
($2{,}058/3{,}792$, Wilson 95\% CI $[52.7,55.9]$), $50.2\%$ for Composed
($956/1{,}903$, $[48.0,52.5]$), and $29.6\%$ for WiCE
($809/2{,}732$, $[27.9,31.4]$). Thus, changes modestly favor the correct label
on RAGTruth, are indistinguishable from chance on Composed, and are net harmful
on WiCE. A changed answer should not be interpreted as a corrected answer.

\subsection{Changing the Reference Model Changes the Pattern}

Table~\ref{tab:crossmodel} shows three additional comparisons in which the
reference model changes while the dataset and prompt format are held fixed.
Several signs change. In particular, VitaminC moves from a large positive
difference with the mini reference to a small negative difference with the
reasoning reference. This sensitivity reinforces the need for a same-model
single-agent control before attributing the observed gaps specifically to
debate.

\begin{table}[t]
\centering
\small
\caption{Sensitivity to the reference model on three datasets.}
\label{tab:crossmodel}
\begin{tabular}{lcccc}
\toprule
Dataset & R1 (mini) & R1 (reasoning) & $\Delta$ (mini) & $\Delta$ (reasoning) \\
\midrule
WiCE & 0.828 & 0.734 & $-0.020$ & $-0.023$ \\
RAGTruth & 0.638 & 0.624 & $+0.020$ & $-0.023$ \\
VitaminC & 0.768 & 0.909 & $+0.085$ & $-0.013$ \\
\bottomrule
\end{tabular}
\end{table}

\section{Mechanism: Recalibration More Than New Evidence}
\label{sec:mechanism}

\subsection{Class-Specific Recall Shifts}

Table~\ref{tab:recall} decomposes the system-level difference into changes in
\grounded{} and \notgrounded{} recall. On VitaminC, most of the gain comes from
a $+18.0$-point recovery in \grounded{} recall while \notgrounded{} recall is
almost unchanged. On RAGTruth, the panel moves in the opposite direction:
\notgrounded{} recall rises by $10.7$ points while \grounded{} recall falls by
$6.7$ points. On SciFact both recalls decline. These directional changes are
consistent with a threshold shift that is beneficial only when it compensates
for the reference system's dominant error.

\begin{table}[t]
\centering
\small
\caption{Class-recall changes from the reference to the panel. The trials match
Table~\ref{tab:main}.}
\label{tab:recall}
\begin{tabular}{lrrrr}
\toprule
Dataset & \grounded{} recall R1$\to$R2 & $\Delta$ grd & $\Delta$ not-grd & $\Delta$ acc. \\
\midrule
Composed & $0.002\to0.038$ & $+0.036$ & $+0.003$ & $+0.020$ \\
HaluEval & $0.841\to0.865$ & $+0.024$ & $-0.013$ & $+0.005$ \\
RAGTruth & $0.837\to0.769$ & $-0.067$ & $+0.107$ & $+0.020$ \\
SciFact & $0.700\to0.629$ & $-0.071$ & $-0.017$ & $-0.044$ \\
VitaminC & $0.584\to0.764$ & $+0.180$ & $-0.011$ & $+0.085$ \\
WiCE & $0.875\to0.795$ & $-0.080$ & $+0.041$ & $-0.020$ \\
\bottomrule
\end{tabular}
\end{table}

\subsection{The Second Round Adds Little Semantic Novelty}

We measure round-two novelty as embedding distance between an agent's
second-round argument and the first-round content available to it, using the
all-MiniLM-L6-v2 representation and the ROSCOE informativeness formulation.
Across approximately $61{,}000$ run--example--agent observations, novelty
relative to all first-round content has median $0.068$, mean $0.108$, standard
deviation $0.117$, fifth percentile $0.000$, and 95th percentile $0.347$.
Novelty relative to peers has median $0.088$; novelty relative to the same
agent's first-round argument has median $0.145$.

The low median does not prove that no second-round argument is useful. It does
show that a typical second-round response is close to content already present
in round one. Together with the recall shifts, this favors recalibration over
evidence expansion as the dominant aggregate mechanism.

\subsection{Reasoning Metrics Weakly Predict Correctness}

We compute eight chain-level ROSCOE metrics on a stratified sample of $3{,}000$
debate rationales. Table~\ref{tab:roscoe} reports correlations with confidence
and correctness. No metric strongly predicts correctness ($|r|\leq0.21$).
Informativeness has the largest magnitude and is negative, suggesting that new
material is more common on difficult or error-prone examples rather than being
a reliable marker of productive deliberation. NLI-based consistency and
hallucination-rate metrics are close to zero.

\begin{table}[t]
\centering
\small
\caption{ROSCOE metric correlations on $n=3{,}000$ debate rationales. ``ns''
marks $p\geq0.05$; repetition is degenerate.}
\label{tab:roscoe}
\begin{tabular}{lrr}
\toprule
Metric & Corr. with confidence & Corr. with correctness \\
\midrule
Informativeness & $-0.148$ & $-0.207$ \\
Coherence & $+0.108$ & $+0.191$ \\
Faithfulness & $+0.139$ & $+0.156$ \\
Consistency (NLI) & $+0.021$ ns & $+0.076$ \\
Hallucination rate (NLI) & $-0.064$ & $-0.053$ \\
Depth & $-0.041$ & $-0.019$ ns \\
Hedging & $-0.011$ ns & $-0.027$ ns \\
Repetition & --- & --- \\
\bottomrule
\end{tabular}
\end{table}

\subsection{Confidence Is Self-Consistent but Poorly Calibrated}

Round-two confidence correlates with correctness at only $+0.10$, but with
round-one confidence at $+0.69$. The AUROC of confidence for predicting
correctness is $0.606$ in round two and $0.535$ in round one. The empirical
reliability curve is non-monotone: examples in the $[0.6,0.7)$ bin are
$62.9\%$ accurate, whereas those in the $[0.9,1.0]$ bin are only approximately
$58\%$ accurate. Debate therefore preserves agents' confidence more strongly
than it aligns confidence with correctness.

\section{Ablations and Routing}
\label{sec:ablations}

\subsection{Speaking Order Is the Largest Tested Debate Knob}

We vary anonymity, temperature, and speaking order. Anonymity changes accuracy
by at most one to two points on Composed, RAGTruth, and WiCE. Temperatures
$\{0.0,0.3,0.7,1.0\}$ yield ranges of at most two points on WiCE and RAGTruth,
with no monotonic trend; these are single runs and do not establish a
temperature effect. Speaking order creates the largest spread of the three
tested knobs, although it remains small on five datasets (Table~\ref{tab:order}).

\begin{table}[t]
\centering
\small
\caption{Panel accuracy across the six speaking orders.}
\label{tab:order}
\begin{tabular}{lrrrrr}
\toprule
Dataset & Mean & SD & Min & Max & Spread \\
\midrule
Composed & 50.50 & 1.18 & 49.00 & 52.50 & 3.50 pp \\
HaluEval & 82.93 & 0.86 & 81.60 & 84.00 & 2.40 pp \\
RAGTruth & 65.77 & 0.63 & 65.20 & 66.80 & 1.60 pp \\
SciFact & 76.93 & 3.58 & 73.20 & 81.20 & 8.00 pp \\
VitaminC & 85.27 & 0.78 & 84.00 & 86.00 & 2.00 pp \\
WiCE & 80.87 & 1.41 & 78.40 & 82.40 & 4.00 pp \\
\bottomrule
\end{tabular}
\end{table}

A conversation-history ablation on WiCE produces a larger effect than these
prompt-level knobs: agents given full prior-round history reach $0.836$
accuracy, compared with $0.788$ when prior history is hidden, a $+4.8$-point
difference. Context construction can therefore matter more than anonymity or
temperature.

\subsection{Order Appears to Interact with Class Bias}

The first speaker tends to pull the panel toward its role's preferred class:
skeptic-first orders favor \notgrounded{}, while advocate-first orders favor
\grounded{}. Across the six datasets, the correlation between reference
accuracy and the system-level difference is $-0.278$. With only six data
points, this is not statistically reliable and should be read as a hypothesis.
A stronger test would regress order-specific recall changes against the
single-agent model's class asymmetry using a same-model baseline.

\subsection{Selective Routing Can Save Calls, but Evidence Is Limited}

On three datasets with retained instance-level outputs, a gradient-boosted tree
uses first-stage verdict, confidence, and disagreement features to predict when
additional debate is harmful. Grouped five-fold out-of-fold evaluation suggests
that debating only the top $5\%$ of WiCE examples changes the measured
difference from $-1.95$ to $+0.58$ points (95\% CI $[+0.23,+0.92]$). In a
separate online study with $50$ examples per dataset, routing saves
$24$--$45\%$ of tokens without a detected accuracy loss. At this sample size,
no per-dataset accuracy difference is significant. We therefore interpret the
online result as preliminary evidence for cost reduction, not accuracy
improvement.

\section{Reflective Prompt Optimization}
\label{sec:gepa}

\subsection{Optimization Question}

The cross-dataset variability could be caused by poorly written persona
prompts. We test this explanation using reflective prompt evolution inspired by
GEPA~\citep{agrawal2025gepa}. A reflector examines failures, proposes a rewrite
to one persona, evaluates the rewritten system, and retains candidates that are
useful on at least part of the development set. The advocate, skeptic, and
domain-expert prompts are optimizable modules; the deterministic aggregator is
frozen.

This distinction creates a direct diagnostic. If persona wording is the main
bottleneck, prompt evolution should improve held-out performance. If the
limitation lies in the shared model, shared evidence, or aggregation rule,
prompt search may find only threshold shifts and highly correlated candidates.

\subsection{Search Variants and Splits}

We evaluate two search organizations:
\begin{itemize}[leftmargin=1.5em]
    \item \textbf{Round-robin:} revisit the three persona modules, maintain a
    Pareto set of candidates, and merge complementary rewrites. This follows
    the main organization of GEPA.
    \item \textbf{Sequential coordinate-ascent control:} optimize one persona
    at a time with greedy parent selection and no merge. We evaluate three
    orders: skeptic-first (seqA), advocate-first (seqB), and domain-first
    (seqC). This control is not the default GEPA algorithm.
\end{itemize}

Initial sequential optimization uses a 24-example development set and a
30-example internal selection set. Final comparisons use a disjoint
120-example held-out set containing 20 examples from each of the six dataset
families and balanced gold labels. A separate paper-aligned split contains 60
training, 120 validation, and 120 held-out examples; its incomplete results are
not used for claims in this paper.

The evaluated family contains a frozen base plus seven alternatives: three
single-persona rewrites, three sequential orders, and one round-robin merged
configuration. All variants use GPT-5.5 Chat and the same fixed aggregator.

\subsection{One Skeptic Rewrite Transfers, but the Evidence Is Exploratory}

Table~\ref{tab:gepa} reports held-out accuracy. The skeptic-first sequential
variant performs best, improving from $53.33\%$ to $59.17\%$. Its paired
McNemar test gives raw $p=0.0156$ and its bootstrap interval is
$[+1.67,+10.0]$ points. A second independent 120-example panel yields a
$+5.00$-point difference (raw $p=0.031$). However, correcting across the seven
alternatives raises the first comparison to approximately $p=0.11$ under a
Bonferroni correction. Repeated base evaluations also vary by approximately
$3.3$ points at $n=120$, comparable to the observed gain. We therefore call the
rewrite promising but exploratory.

\begin{table}[t]
\centering
\small
\caption{Held-out accuracy for the frozen base and seven prompt alternatives
($n=120$). Only seqA has an uncorrected interval excluding zero.}
\label{tab:gepa}
\begin{tabular}{lrr}
\toprule
Configuration & Accuracy & Difference from base \\
\midrule
Base & 53.33 & --- \\
Single skeptic rewrite & 56.67 & $+3.34$ pp \\
Single advocate rewrite & 55.00 & $+1.67$ pp \\
Single domain rewrite & 55.00 & $+1.67$ pp \\
seqA: skeptic first & 59.17 & $+5.83$ pp \\
seqB: advocate first & 54.17 & $+0.84$ pp \\
seqC: domain first & 50.83 & $-2.50$ pp \\
Round-robin merge & 57.50 & $+4.17$ pp \\
\bottomrule
\end{tabular}
\end{table}

The seqA rewrite changes two load-bearing instructions. It explicitly protects
valid paraphrase, coreference, and direct inference from over-rejection, and it
maps missing support to \notgrounded{} rather than \abstain{}. On held-out
data, seqA fixes seven examples and breaks none relative to the frozen base; all
seven fixes are cases where the base abstains and seqA commits the correct
binary label. This is a clean threshold change, not new evidence acquisition.

\subsection{Small Development Sets Hide Overfitting}

The 24-example development set cannot distinguish seqA from seqC: both score
$66.7\%$ ($16/24$). On the shared 120-example held-out set, they differ by
$8.34$ points (Table~\ref{tab:overfit}). SeqC optimizes the domain prompt and
never modifies the skeptic; its development gain does not transfer. This case
illustrates why a prompt optimizer must select on validation data and reserve a
separate held-out set.

\begin{table}[t]
\centering
\small
\caption{Development--held-out divergence for two sequential orders.}
\label{tab:overfit}
\begin{tabular}{llrrr}
\toprule
Config. & Edited role & Dev ($n=24$) & Held out ($n=120$) & $\Delta$ vs. base \\
\midrule
Base & --- & 62.5 & 53.33 & --- \\
seqA & Skeptic & 66.7 & 59.17 & $+5.83$ pp \\
seqC & Domain expert & 66.7 & 50.83 & $-2.50$ pp \\
\bottomrule
\end{tabular}
\end{table}

\subsection{The Evaluated Prompt Family Has No Ensemble Headroom}

A per-example oracle over the seven alternatives equals seqA exactly: every
item solved by any alternative is already solved by seqA. Consequently, no
vote, weight, or stack over this evaluated family can exceed seqA on the
held-out set. This does not prove that prompt ensembles are generally useless;
it shows that these candidates lack complementary errors.

More importantly, $49$ of $120$ held-out examples are wrong under the base and
every prompt alternative. Of these persistent errors, $40$ are \grounded{}
claims predicted as \notgrounded{}. The search therefore finds a small
recalibration but does not create a new verification capability.

\section{Correlated Errors and the Structural Ceiling}
\label{sec:correlation}

We compute the $\phi$ correlation between binary persona error indicators,
pooling the six speaking orders within each dataset and using a cluster
bootstrap by example. Table~\ref{tab:corr} shows two regimes. The advocate and
domain expert remain strongly coupled on every dataset ($0.671$--$0.927$). The
skeptic is also highly coupled on Composed and HaluEval, but decouples from the
advocate on WiCE ($0.288$) and SciFact ($0.375$).

\begin{table}[t]
\centering
\small
\caption{Pairwise persona error correlation $\phi$ with 95\% cluster-bootstrap
intervals. The pooled row should not replace the per-dataset pattern.}
\label{tab:corr}
\begin{tabular}{lccc}
\toprule
Dataset & Skeptic--advocate & Skeptic--domain & Advocate--domain \\
\midrule
Composed & .983 $[.972,.993]$ & .867 $[.826,.903]$ & .867 $[.827,.903]$ \\
HaluEval & .790 $[.716,.856]$ & .853 $[.793,.907]$ & .925 $[.889,.957]$ \\
RAGTruth & .521 $[.464,.580]$ & .766 $[.729,.804]$ & .762 $[.726,.801]$ \\
SciFact & .375 $[.292,.466]$ & .532 $[.470,.598]$ & .888 $[.853,.924]$ \\
VitaminC & .527 $[.406,.627]$ & .733 $[.656,.802]$ & .671 $[.586,.746]$ \\
WiCE & .288 $[.235,.340]$ & .405 $[.364,.452]$ & .927 $[.907,.946]$ \\
\midrule
Pooled & .516 $[.485,.547]$ & .645 $[.621,.669]$ & .858 $[.843,.872]$ \\
\bottomrule
\end{tabular}
\end{table}

The result is more specific than saying that all personas always fail as one
block. The skeptic contributes diversity on the harder entailment datasets,
but the advocate and domain expert remain redundant. Because majority voting
executes before the skeptic veto, a correlated advocate--domain pair can
determine the output even where the skeptic is informative. Prompt optimization
cannot change that ordering. The structural ceiling is therefore the
interaction of shared evidence, correlated model errors, and a frozen
aggregation rule.

\section{Discussion}

\paragraph{Debate as recalibration.}
Three observations support the recalibration account. First, typical
round-two arguments have low novelty. Second, the largest performance changes
are directional class-recall shifts. Third, the useful GEPA rewrite changes
the skeptic's reject/abstain boundary and rescues seven abstentions without
adding evidence. Debate may still surface overlooked details on individual
examples, but evidence expansion is not the dominant aggregate explanation.

\paragraph{Why task dependence matters.}
A threshold shift helps when it counters a dataset's dominant error and harms
when it moves an already adequate boundary. VitaminC benefits from recovering
under-predicted \grounded{} examples; RAGTruth benefits from increasing
\notgrounded{} recall; SciFact loses recall in both classes. A single debate
policy is therefore unlikely to be uniformly optimal across domains.

\paragraph{Prompt optimization as a diagnostic.}
GEPA-style search is valuable here even without a robust headline gain. It
identifies a specific skeptic instruction that changes behavior, reveals severe
development-set overfitting, and demonstrates that the evaluated prompt family
has no complementary ensemble errors. The null headroom result localizes the
next intervention: optimize the aggregator or introduce genuinely independent
information sources.

\paragraph{Design implications.}
Future groundedness panels should consider: (i) heterogeneous models with
measured error complementarity; (ii) different retrieved evidence windows or
tools for different agents; (iii) an aggregation rule that uses calibrated
role reliability rather than a fixed majority; and (iv) an explicit abstention
policy trained on evidence sufficiency. These are hypotheses motivated by the
present analysis, not results demonstrated here.

\section{Limitations}

\paragraph{The main comparison does not isolate debate.}
The single-agent reference uses GPT-5 mini, whereas the panel uses GPT-5.5 Chat.
The reported cross-dataset differences therefore combine model and system
effects. A causal debate ablation must compare one GPT-5.5 Chat judge with the
GPT-5.5 Chat panel under identical prompts, decoding, and evidence. This is the
largest limitation of the current study.

\paragraph{Prompt search is exploratory.}
The primary GEPA study uses a small development set, one main optimization
seed, seven alternatives, and a 120-example held-out set. The strongest raw
result does not survive Bonferroni correction, and model nondeterminism is of
the same order as the observed gain. The replication is encouraging but does
not replace a multi-seed, validation-selected protocol.

\paragraph{The optimizer cannot modify the aggregator.}
Our intervention changes persona prompts while freezing the deterministic
meta-judge. The conclusion is therefore limited to prompt-level optimization
of this panel. A jointly optimized or learned aggregator may behave
differently.

\paragraph{Dataset conversion changes the original tasks.}
We map heterogeneous public labels into a balanced binary groundedness task and
construct a four-claim WiCE conjunction. These transformations enable a shared
evaluation but may remove natural base rates and dataset-specific distinctions.
Results should not be read as leaderboard scores on the original tasks.

\paragraph{Limited diversity.}
All personas share one model and evidence pool. We do not test heterogeneous
models, independent retrieval, tools, more than three agents, or more than two
rounds. The structural explanation applies to the evaluated homogeneous panel,
not every possible multi-agent system.

\paragraph{Cost measurement.}
The panel makes six LLM calls per example, but wall-clock latency was not logged
consistently. Routing results use token savings and small online samples; a
complete deployment comparison should report latency, monetary cost, and
quality under matched service conditions.

\section{Conclusion}

Homogeneous multi-agent debate is not a uniformly reliable groundedness judge.
Across six benchmarks, its system-level difference from a fixed reference is
positive, negative, or inconclusive depending on the task. Within the panel,
low second-round novelty, directional recall shifts, weak confidence
calibration, and correlated persona errors indicate that discussion primarily
recalibrates existing judgments. Reflective prompt optimization discovers one
promising skeptic rewrite but leaves a large set of persistent errors and no
ensemble headroom among the tested prompts.

The practical lesson is not simply to add more agents or rewrite their
personas. When agents share a model and evidence pool, improvements depend on
whether their errors are genuinely complementary and whether the aggregator
can exploit that complementarity. Stronger groundedness judges will likely
require independent evidence or model diversity together with calibrated
aggregation and abstention.

\appendix

\section{Reproducibility Checklist}

Before public release, the accompanying artifact should record:
\begin{itemize}[leftmargin=1.5em]
    \item exact GPT-5.5 Chat and GPT-5 mini API model identifiers and access
    dates;
    \item decoding parameters, random seeds, retry policy, and parsing rules;
    \item the full three persona prompts and every optimized prompt;
    \item the deterministic aggregation implementation;
    \item dataset versions, original splits, binary-label mappings, sampling
    seeds, and identifiers for the 1,950 examples;
    \item cluster-bootstrap, Wilson-interval, and McNemar-test code;
    \item all per-example verdicts, confidences, speaking orders, and prompt
    configuration identifiers needed to reproduce the tables.
\end{itemize}

\section{Claim-to-Evidence Audit}

Table~\ref{tab:audit} records which results support each central claim and what
the result does not establish.

\begin{table}[h]
\centering
\small
\caption{Scope audit for the paper's principal claims.}
\label{tab:audit}
\begin{tabular}{p{3.2cm}p{5.0cm}p{5.0cm}}
\toprule
Claim & Supporting result & Not established \\
\midrule
Performance is task-dependent & Table~\ref{tab:main}; intervals vary in sign &
Pure causal effect of debate \\
Panel behavior is largely recalibration & Recall shifts, low novelty, GEPA's
seven rescued abstentions & No individual example ever gains new insight \\
Confidence is unreliable & AUROC $0.606$ at round two; non-monotone bins &
All confidence-calibration methods will fail \\
Prompt family has limited diversity & Oracle over seven alternatives equals
seqA; $49/120$ wrong for all & All possible prompts or optimizers lack headroom \\
Errors are structurally coupled & Table~\ref{tab:corr}; frozen majority rule &
Heterogeneous models or evidence will also fail \\
\bottomrule
\end{tabular}
\end{table}


\begin{thebibliography}{99}

\bibitem[Agrawal et~al.(2025)Agrawal, Tan, Soylu, Ziems, Khare,
Opsahl-Ong, Singhvi, Shandilya, Ryan, Jiang, Potts, Sen, Dimakis, Stoica,
Klein, Zaharia, and Khattab]{agrawal2025gepa}
Lakshya A. Agrawal, Shangyin Tan, Dilara Soylu, Noah Ziems, Rishi Khare,
Krista Opsahl-Ong, Arnav Singhvi, Herumb Shandilya, Michael J. Ryan, Meng
Jiang, Christopher Potts, Koushik Sen, Alexandros G. Dimakis, Ion Stoica, Dan
Klein, Matei Zaharia, and Omar Khattab. 2025.
\newblock GEPA: Reflective prompt evolution can outperform reinforcement
learning.
\newblock \emph{arXiv preprint arXiv:2507.19457}.

\bibitem[Chan et~al.(2023)Chan, Chen, Su, Yu, Xue, Zhang, Fu, and
Liu]{chan2023chateval}
Chi-Min Chan, Weize Chen, Yusheng Su, Jianxuan Yu, Wei Xue, Shanghang Zhang,
Jie Fu, and Zhiyuan Liu. 2023.
\newblock ChatEval: Towards better LLM-based evaluators through multi-agent
debate.
\newblock \emph{arXiv preprint arXiv:2308.07201}.

\bibitem[Chen et~al.(2026)Chen, Zhao, Zhong, Wang, Shi, and
Zheng]{chen2026dlsg}
Yisong Chen, Chuqing Zhao, Yishan Zhong, Ziyu Wang, Jiazhao Shi, and Wenjia
Zheng. 2026.
\newblock Applying the Deep Learning--Sector--Governance (DLSG) framework to
the U.S. healthcare system: Opportunities, deployment pathways, and
policy-aligned evaluation.
\newblock \emph{Journal of Technology Innovation and Society}, 4:1--18.
\newblock \href{https://doi.org/10.63646/j.jtis.2026.040101}{doi:10.63646/j.jtis.2026.040101}.

\bibitem[Chen et~al.(2025)Chen, Gu, and Ye]{chen2025design}
Rensi Chen, Boping Gu, and Ziyi Ye. 2025.
\newblock Design and implementation of big data-driven business intelligence
analytics system.
\newblock In \emph{Image Processing, Electronics and Computers}, pages
1219--1227. IOS Press.

\bibitem[Du et~al.(2023)Du, Li, Torralba, Tenenbaum, and Mordatch]{du2023debate}
Yilun Du, Shuang Li, Antonio Torralba, Joshua B. Tenenbaum, and Igor Mordatch.
2023.
\newblock Improving factuality and reasoning in language models through
multiagent debate.
\newblock \emph{arXiv preprint arXiv:2305.14325}.

\bibitem[Golovneva et~al.(2022)Golovneva, Chen, Poff, Corredor, Zettlemoyer,
Fazel-Zarandi, and Celikyilmaz]{golovneva2022roscoe}
Olga Golovneva, Moya Chen, Spencer Poff, Martin Corredor, Luke Zettlemoyer,
Maryam Fazel-Zarandi, and Asli Celikyilmaz. 2022.
\newblock ROSCOE: A suite of metrics for scoring step-by-step reasoning.
\newblock \emph{arXiv preprint arXiv:2212.07919}.

\bibitem[Kamoi et~al.(2023)Kamoi, Goyal, Rodriguez, and Durrett]{kamoi2023wice}
Ryo Kamoi, Tanya Goyal, Juan Diego Rodriguez, and Greg Durrett. 2023.
\newblock WiCE: Real-world entailment for claims in Wikipedia.
\newblock In \emph{Proceedings of EMNLP}, pages 7561--7583.

\bibitem[Liang et~al.(2025)Liang, Wang, Li, Zhu, Jiang, Gong, and
Wang]{liang2025graphrag}
Jiacheng Liang, Yuhui Wang, Changjiang Li, Rongyi Zhu, Tanqiu Jiang, Neil Gong,
and Ting Wang. 2025.
\newblock GraphRAG under fire.
\newblock \emph{arXiv preprint arXiv:2501.14050}.

\bibitem[Li et~al.(2023)Li, Cheng, Zhao, Nie, and Wen]{li2023halueval}
Junyi Li, Xiaoxue Cheng, Wayne Xin Zhao, Jian-Yun Nie, and Ji-Rong Wen. 2023.
\newblock HaluEval: A large-scale hallucination evaluation benchmark for large
language models.
\newblock In \emph{Proceedings of EMNLP}, pages 6449--6464.

\bibitem[Liu et~al.(2023)Liu, Iter, Xu, Wang, Xu, and Zhu]{liu2023geval}
Yang Liu, Dan Iter, Yichong Xu, Shuohang Wang, Ruochen Xu, and Chenguang Zhu.
2023.
\newblock G-Eval: NLG evaluation using GPT-4 with better human alignment.
\newblock In \emph{Proceedings of EMNLP}.

\bibitem[Niu et~al.(2024)Niu, Wu, Zhu, Xu, Shum, Zhong, Song, and
Zhang]{niu2024ragtruth}
Cheng Niu, Yangneng Wu, Jiadong Zhu, Siliang Xu, KaShun Shum, Randy Zhong,
Jayanth G. D. Song, and Tong Zhang. 2024.
\newblock RAGTruth: A hallucination corpus for developing trustworthy
retrieval-augmented language models.
\newblock In \emph{Proceedings of ACL}.

\bibitem[Schuster et~al.(2021)Schuster, Fisch, and
Barzilay]{schuster2021vitaminc}
Tal Schuster, Adam Fisch, and Regina Barzilay. 2021.
\newblock Get your Vitamin C! Robust fact verification with contrastive
evidence.
\newblock In \emph{Proceedings of NAACL}, pages 624--643.

\bibitem[Song et~al.(2026)Song, Zheng, Chen, Wang, Shi, and Chen]{song2026deep}
Yun Song, Wenjia Zheng, Tiedan Chen, Ziyu Wang, Jiazhao Shi, and Yisong Chen.
2026.
\newblock Deep neural network architectures for electrocardiogram
classification: A comprehensive evaluation.
\newblock \emph{arXiv preprint arXiv:2602.17701}.

\bibitem[Tian et~al.(2024)Tian, Wang, Zhao, and Ding]{tian2024mmrec}
Jiahao Tian, Zhenkai Wang, Jinman Zhao, and Zhicheng Ding. 2024.
\newblock MMRec: LLM based multi-modal recommender system.
\newblock In \emph{2024 19th International Workshop on Semantic and Social
Media Adaptation \& Personalization (SMAP)}, pages 105--110. IEEE.

\bibitem[Wadden et~al.(2020)Wadden, Lin, Lo, Wang, van Zuylen, Cohan, and
Hajishirzi]{wadden2020scifact}
David Wadden, Shanchuan Lin, Kyle Lo, Lucy Lu Wang, Madeleine van Zuylen,
Arman Cohan, and Hannaneh Hajishirzi. 2020.
\newblock Fact or fiction: Verifying scientific claims.
\newblock In \emph{Proceedings of EMNLP}, pages 7534--7550.

\bibitem[Wang et~al.(2025a)Wang, Li, Chen, Liang, and
Wang]{wang2025reasoningretrievalstudyanswer}
Yuhui Wang, Changjiang Li, Guangke Chen, Jiacheng Liang, and Ting Wang. 2025a.
\newblock Reasoning or retrieval? A study of answer attribution on large
reasoning models.
\newblock \emph{arXiv preprint arXiv:2509.24156}.

\bibitem[Wang and Tian(2025)Wang and Tian]{wang2025dlrrec}
Zhenkai Wang and Jiahao Tian. 2025.
\newblock DLRREC: Denoising latent representations via multi-modal knowledge
fusion in deep recommender systems.
\newblock In \emph{Proceedings of the 2025 9th International Conference on
Computer Science and Artificial Intelligence}, pages 575--581.

\bibitem[Liu et~al.(2025)Liu, Yang, and Xia]{liu2025research}
Kuangcong Liu, Shini Yang, and Jiayi Xia. 2025.
\newblock Research and practice of advertisement recommendation algorithm based
on graph neural network.
\newblock In \emph{Proceedings of the 2nd International Symposium on
Integrated Circuit Design and Integrated Systems}, pages 210--215.

\bibitem[Zheng et~al.(2023)Zheng, Chiang, Sheng, Zhuang, Wu, Zhuang, Lin, Li,
Li, Xing, Zhang, Gonzalez, and Stoica]{zheng2023judging}
Lianmin Zheng, Wei-Lin Chiang, Ying Sheng, Siyuan Zhuang, Zhanghao Wu,
Yonghao Zhuang, Zi Lin, Zhuohan Li, Dacheng Li, Eric P. Xing, Hao Zhang,
Joseph E. Gonzalez, and Ion Stoica. 2023.
\newblock Judging LLM-as-a-judge with MT-Bench and Chatbot Arena.
\newblock \emph{arXiv preprint arXiv:2306.05685}.

\end{thebibliography}
\end{document}